\documentclass[11pt,a4paper]{article}
\usepackage[hyperref]{acl2021}
\usepackage{times}
\usepackage{latexsym}

\usepackage{graphicx}
\usepackage{microtype}
\usepackage{makecell}
\usepackage{multirow}

\usepackage{amsmath}
\DeclareMathOperator*{\argmax}{argmax}

\usepackage{xspace}
\newcommand*{\eg}{e.g.\@\xspace}

\aclfinalcopy 
\def\aclpaperid{3233} 

\title{Who Blames or Endorses Whom? \\Entity-to-Entity Directed Sentiment Extraction in News Text}

\author{Kunwoo Park\\
  Soongsil University\thanks{This work was done while the first author was a postdoctoral researcher at UCLA. }
  \\
  {\small \texttt{kunwoo.park@ssu.ac.kr}} \\\And
  Zhufeng Pan\\
  UCLA\\
  {\small \texttt{panzhufeng@cs.ucla.edu}} \\
  \And
  Jungseock Joo\\
  UCLA\\
  {\small \texttt{jjoo@comm.ucla.edu}}}

\begin{document}
\maketitle
\begin{abstract}
Understanding who blames or supports whom in news text is a critical research question in computational social science. Traditional methods and datasets for sentiment analysis are, however, not suitable for the domain of political text as they do not consider the direction of sentiments expressed between entities. 
In this paper, we propose a novel NLP task of identifying directed sentiment relationship between political entities from a given news document, which we call \textit{directed sentiment extraction}. From a million-scale news corpus, we construct a dataset of news sentences where sentiment relations of political entities are manually annotated. We present a simple but effective approach for utilizing a pretrained transformer, which infers the target class by predicting multiple question-answering tasks and combining the outcomes. We demonstrate the utility of our proposed method for social science research questions by analyzing positive and negative opinions between political entities in two major events: 2016 U.S. presidential election and COVID-19. The newly proposed problem, data, and method will facilitate future studies on interdisciplinary NLP methods and applications.\footnote{\url{https://github.com/bywords/directed\_sentiment\_analysis}}
\end{abstract}

\section{Introduction}
Sentiment analysis is a useful technique for opinion mining from text data. Most existing work either focuses on sentence-level classification~\citep{van2018semeval, zampieri2019predicting}, or aims to detect the sentiment polarity towards specific targets~\citep{pontiki2016semeval, cortis2017semeval}.  
These approaches typically do not distinguish sources and targets of the sentiment. They mainly use user-generated content (UGC), such as tweets or restaurant reviews from Yelp, which assumes each user (account holder) is the source of the sentiment, and that the target is also clearly defined or easily identifiable (\eg, the restaurant). 

This assumption does not hold for political news analysis where a large number of political actors blame or support each other in news articles. The key interest for political sentiment analysis is to identify ``who'' blame or support ``whom''~\citep{balahur2009rethinking} rather than simply assigning a global sentiment polarity to a given document or sentence. For example, from a sentence like ``X supported Y for criticizing Z,'' we can infer X is positive toward Y and both X and Y are negative toward Z. However, existing sentiment analysis methods are not suitable to detect such sentiment relationships between entities. 

\begin{figure}[t]
\centering
\includegraphics[width=.8\linewidth]{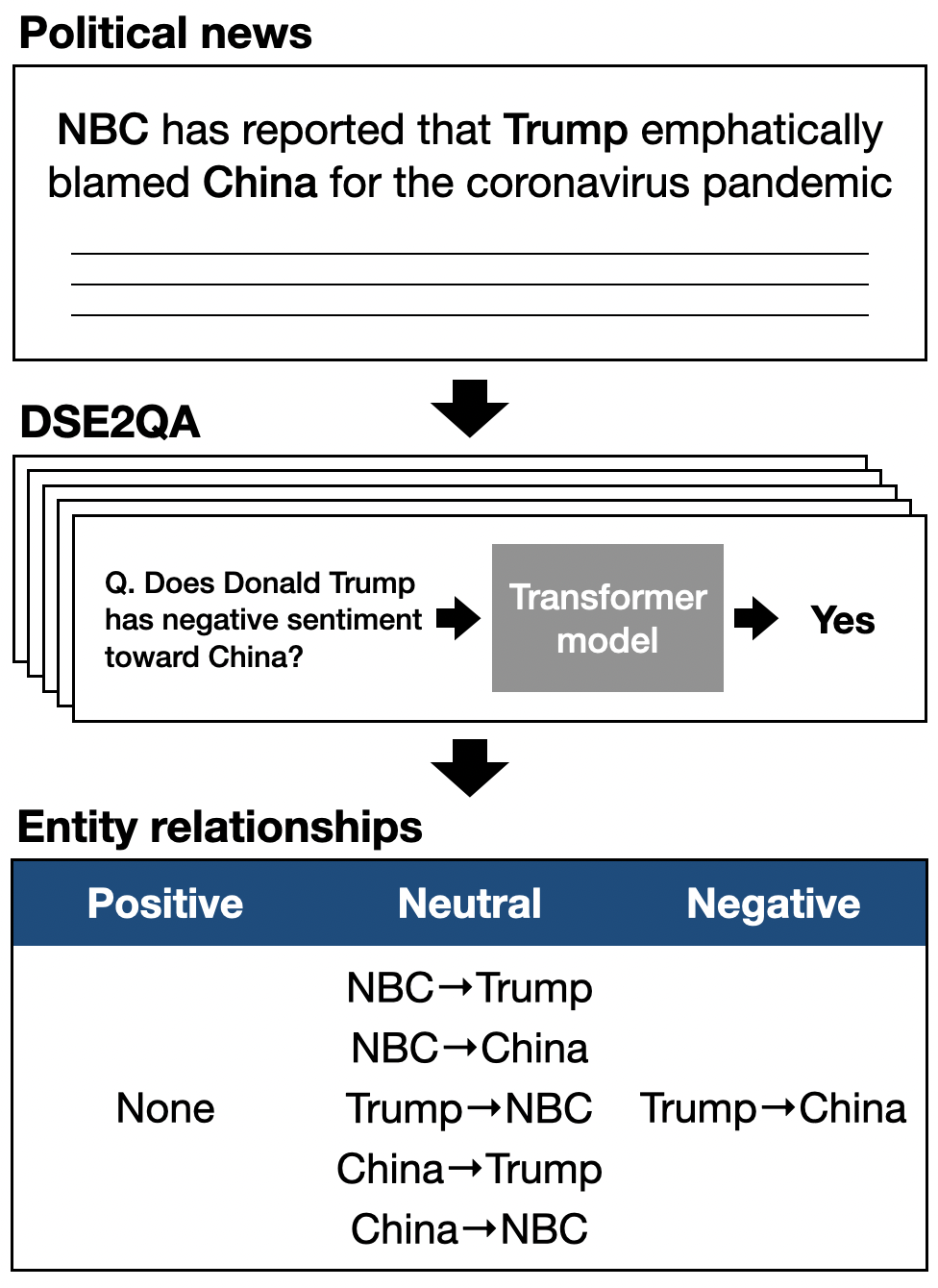}
\caption{Overview of the directed sentiment extraction}\vspace{-3mm}
\label{fig:overview}
\end{figure}

This paper proposes a new NLP task of identifying directed sentiment relationships from a political entity to another in news articles: \textit{directed sentiment extraction}. For this task, we introduce a newly annotated corpus from a million-scale news dataset. As demonstrated in Figure~\ref{fig:overview}, we transform directed sentiment extraction to multiple question-answering tasks (\textbf{DSE2QA}) and combine their predictions for making a final prediction. Evaluation results show it outperforms state-of-the-art classification approaches, such as a fine-tuned RoBERT classification model. Going further, we demonstrate the utility of our method through two case studies on how news media in the U.S. portrayed relationships between political entities differently amid the US election and COVID-19 pandemic. 
The analysis reveals that the left-leaning media present Donald Trump more as the target of blame while the right-leaning media present news stories blaming other entities. This study not only makes a contribution to the NLP community by defining a new problem and approach but also adds the empirical understanding of media bias to the social science community.

\section{Related Work}\label{sec:related-work}
\subsection{Sentiment Analysis in Media}

Sentiment analysis or polarity detection aims at deciding whether a given text contains positive or negative sentiment~\cite{liu2012sentiment} or quantifying the degree of sentiments embedded in a text~\cite{gilbert2014vader}. 
Previous studies have tackled the problem as a classification task~\cite{maas2011learning} and 
applied the trained model to infer sentiments embedded in various web and social data~\cite{park2018positivity}. 
Recently, transformer-based models have shown high performance in sentiment classification~\cite{devlin2018bert} and aspect-based sentiment analysis~\cite{sun2019utilizing}. 

Measuring sentiment or tone of political text in news media is a widely  used method in computational social science~\cite{young2012affective}.  A stream of work has used social media posts to estimate public opinions about political actors and predict the outcomes of future events by large scale sentiment analysis~\cite{o2010tweets,ceron2014every,wang2012system}, while some studies further extend to nonverbal or multimodal dimensions~\cite{joo2014visual, won2017protest, chen2020understanding}. 

\subsection{Stance Detection}
Stance detection is a relevant NLP problem that aims to predict a claim's stance on reference text~\cite{augenstein2016stance} or to infer social media users' view toward an issue~\cite{darwish2020unsupervised}. Many studies tried to advance the deep learning-based models~\cite{mohtarami2018automatic}, for example, by modeling text with a hierarchical architecture~\cite{sun2018stance,yoon2019detecting}. 
Unlike stance detection, this study aims at understanding an entity's sentiment toward another entity, both of which appear in the same sentence.

\subsection{Relation Extraction}

Our target problem is also relevant to relation extraction, which is a task of extracting structured relationships between entities from unstructured text. While the early literature relied on feature-based methods~\cite{zelenko2003kernel}, recent methods actively utilize neural methods~\cite{lin2016neural}; for example, a study proposed a neural model that jointly learns to perform entity recognition and relation extraction~\cite{bekoulis2018joint}. Most recently, a study tested the use of the pretrained transformer-based language model for relation extraction~\cite{zhang2020can}. 

Despite its similarity to directed sentiment extraction, most of the existing datasets only consider \emph{explicit} entity relationships such as \textsc{employee\_of} and \textsc{city\_of\_residence}~\cite{zhang2017position}. Understanding sentiment relationships between political entities is more challenging as their sentiment is usually hidden in text.

\section{Problem and Dataset}\label{sec:dataset}

In this section, we introduce our problem formulation and explain the process of our dataset construction and annotation. 

\subsection{Target Problem}

Given a sentence $s$ that contains two entities $p$ and $q$, the \textit{directed sentiment extraction} problem aims to detect the sentiment relation from $p$ to $q$ among five classes: neutral, $p$ holds a positive or negative opinion towards $q$, and the reverse direction. For example, in the given sentence in Figure~\ref{fig:overview}, the model should infer that Trump is the source of the negative sentiment toward China, the target. Existing approaches for sentiment analysis cannot be easily adapted to the task, as existing methods aim to detect polarity embedded in a text (sentiment classification), for a specific target (targeted), or with regard to an aspect (aspect-based). These problem setups do not consider the source and target of the sentiment at a time, which cannot identify directed sentiment relationships between political entities. In the following, we introduce a new annotated corpus for the problem.

\subsection{Data Collection}
To construct our dataset, we used news articles from the Real News corpus~\cite{zellers2019defending}, which consists of 32,797,763 real news stories in English published by various outlets over multiple years. Among them, we used 7,127,692 news articles shared by news media that are verified as trustworthy by Media Bias/Fact Check~\cite{mediabiasfactcheck}.
After splitting each article into multiple sentences, we only took sentences with two or more entities using the named entity recognition tool in Spacy\footnote{https://spacy.io}. 
To focus on the relationships between political entities, we considered named entities identified as people, countries/political groups, organizations, or cities/states\footnote{https://spacy.io/api/annotation\#named-entities}.

Since most entity relationships in regular sentences are presumably neutral, we took two approaches for sampling sentences that cover diverse relationships: (i) dictionary-based and (ii) random selection approaches. The dictionary-based approach filters in sentences containing positive or negative keywords from a pre-defined dictionary. Starting from the blame-related keywords~\cite{liang2019blames}, we extended the dictionary by adding their synonyms and antonyms. While this method is  effective in sampling from an unbalanced dataset, it excludes sentences that do not explicitly mention a blame or support keyword. Thus, we also randomly drew sentences to improve the dataset coverage. 

\begin{table}
\centering
\begin{tabular}{|cr|}
\hline 
Class & Count\\\hline
\makecell{Neutral} & 10,604\\
\makecell{Positive ($p \rightarrow q$)} & 1,656\\
\makecell{Positive ($p \leftarrow q$)} & 327\\
\makecell{Negative ($p \rightarrow q$)} & 3,163\\
\makecell{Negative ($p \leftarrow q$)} & 478\\\hline
\makecell{Total} & 16,228\\ 
\hline
\end{tabular}
\caption{Dataset statistics}
\label{table:data-stat}
\end{table}

\subsection{Crowdsourced Annotation}
We used Amazon Mechanical Turk (AMT) to annotate each sentence. The annotation task asked workers to identify what sentiment does an entity holds toward another in a given sentence.  We instructed them to annotate a sentence based only on the sentiment expressed within the sentence, without relying on prior knowledge. There are five options to choose from, neutral, positive ($p \rightarrow q$), positive ($q\rightarrow p$), negative ($p \rightarrow q$), and negative ($q \rightarrow p$). $p$ is the preceding entity of $q$, and the arrow indicates sentiment direction between the two entities. The detailed instruction used for educating annotator is presented in Table~\ref{tab:instruction} in Appendix.

We hired five workers to annotate each sentence. For ensuring high-quality responses, we only allowed workers to participate in the task when they had at least a 70\% acceptance rate for more than 1,000 previous annotations. After completing the initial round of annotation, we filtered out unreliable responses completed within one second or responses by workers who did not pass test questions. We designated workers with at least one unreliable response as untrustworthy and discarded all the other responses submitted by the untrustworthy workers. We repeated the annotation task for the discarded answers until we have five annotations for every sentence. The final set of annotations indicates a Fleiss' kappa value of 0.26, indicating an acceptable level of agreement among annotators. The level of reliability is comparable to studies using subjective text annotations such as hate speech annotation~\cite{ross2017measuring}, subjectivity~\cite{abdul2011subjectivity}, and sentiment analysis~\cite{park2018positivity}.
By aggregating five responses for each sentence by a majority vote, we obtained the final dataset of 16,228 sentences of which sentiment direction is annotated, as shown in Table~\ref{table:data-stat}. While keeping the label distribution almost identical, we split the dataset into 13144, 1461, and 1623 instances for train, validation, and test set, respectively.

\begin{figure*}[t]
    \centering
    \includegraphics[width=0.9\textwidth]{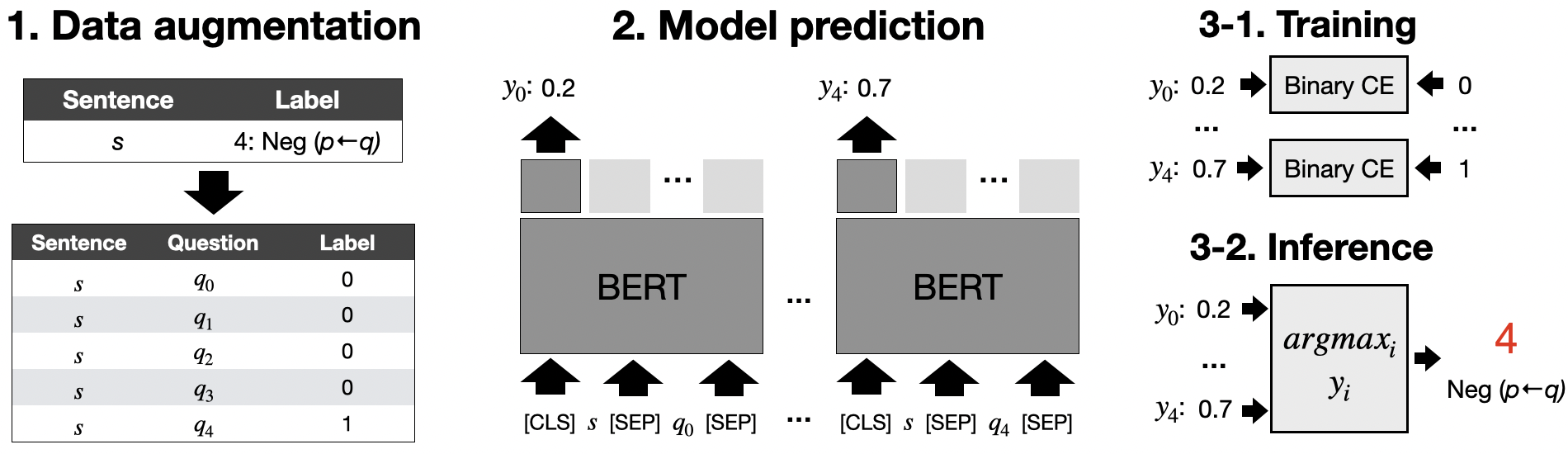}
    \caption{Detailed illustration of the DSE2QA approach}
    \vspace{-1mm}
    \label{fig:dse2qa}
    
\end{figure*}

\section{Methods}\label{sec:methodology}

This section presents methods for addressing the problem of directed sentiment extraction. We propose a novel approach for solving it by employing augmented inputs in BERT-like transformer models and compare it with classification approaches.

\subsection{Classification Approaches}

Following the standard in using classification setups for (undirected) sentiment detection~\cite{liu2012sentiment,devlin2018bert}, we construct classification models that predict scores of each sentiment for the directed sentiment extraction.

\subsubsection{Existing Methods}

We first construct previous approaches proposed for blame relationship detection. \citet{liang2019blames} proposed three neural models that predict the classes of blame relationships in a given text: $p\rightarrow q$, $p\leftarrow q$, and no relationship. We extend their approaches to the five classes of directed sentiment in our dataset.

\emph{Entity prior} model exploits the prior knowledge on political entities by using the representation of two entities using a pre-trained word embedding: $\mathbf{e}_{p}$ and $\mathbf{e}_{q}$. After concatenation, the two vectors are fed into a fully-connected network with a ReLU hidden layer to make a final prediction. \emph{Context} model utilizes the context of the target sentence where two entities appear. After replacing $p$ and $q$ with special tokens (\textsc{[Ent1]} and \textsc{[Ent2]}) respectively, the model encodes the input text through a bidirectional LSTM and extracts the representation corresponding to the two entities: $\mathbf{lstm}_{p}$ and $\mathbf{lstm}_{q}$. The representation gets fed into a fully-connected network. \emph{Combined} model utilizes the concatenated representation of the outputs of the entity prior and context model: $\mathbf{e}_{p}$, $\mathbf{e}_{q}$, $\mathbf{lstm}_{p}$, and $\mathbf{lstm}_{q}$. Then, the vector is fed into a fully-connected network. We train the existing models by minimizing the cross-entropy loss.

\subsubsection{Fine-Tuning a Pretrained Transformer}

Fine-tuning a BERT-like pretrained transformer has shown significant performance in many downstream tasks~\cite{devlin2018bert}.
We also evaluate the performance of a fine-tuned transformer. In particular, after replacing the tokens corresponding to $p$ and $q$ with \textsc{[Ent1]} and \textsc{[Ent2]}, we train a classification model that predicts the five-class output based on the representation of \textsc{[CLS]}\footnote{$<$\textsc{s}$>$ in RoBERTa}. We use the RoBERTa base model~\cite{liu2019roberta} as backbone, and we refer to the classification model as \textit{RoBERTa} in evaluation experiments. The model is trained to minimize the cross-entropy loss.

\begin{table*}[ht]
    \centering
    \begin{tabular}{|c|l|l|}
    \hline
        Index & \makecell{Complete questions} & \makecell{Pseudo questions} \\\hline
        $q_0$ & Do [Ent1] and [Ent2] have neutral sentiment toward each other? & [Ent1] - [Ent2] - neutral\\
        $q_1$ & Does [Ent1] has positive sentiment toward [Ent2]? & [Ent1] - [Ent2] - positive\\
        $q_2$ & Does [Ent2] has positive sentiment toward [Ent1]? & [Ent2] - [Ent1] - positive\\
        $q_3$ & Does [Ent1] has negative sentiment toward [Ent2]? & [Ent1] - [Ent2] - negative\\
        $q_4$ & Does [Ent2] has negative sentiment toward [Ent1]? & [Ent2] - [Ent1] - negative\\\hline
    \end{tabular}
    \caption{Auxiliary questions according to the target label}
    \vspace{-2mm}
    \label{tab:aux_question}
\end{table*}

\subsection{Proposed Approach: Directed Sentiment Extraction to Question-Answering}

BERT-like transformer models are pretrained using two inputs including a separator token\footnote{\textsc{[CLS]} in BERT, $<$\textsc{/s}$><$\textsc{/s}$>$ in RoBERTa} with varying training objectives. The input configuration allows the model to be successfully transferred to the tasks using an auxiliary input, such as sentence pair classification (sentence 1 and sentence 2) and question answering (reference and question). Inspired by the recent achievements using auxiliary inputs~\cite{sun2019utilizing,cohen2020relation}, we propose a simple but effective approach of tackling the directed sentiment extraction problem, which we call \textbf{DSE2QA}; we transform the sentiment extraction task into the sub-tasks aiming for answering yes/no questions on whether a target sentiment is embedded in the text. The basic idea is we inquire an intelligent machine who can answer yes/no questions on whether a target sentiment exists and then combine the answers corresponding to the each sentiment class for making a final guess. Figure~\ref{fig:dse2qa} presents the overall framework, which we elaborate on each step in the following. Technically, taking auxiliary input in BERT-like transformers enables implementing the intelligent machine by making a different prediction with the same sentence input, according to the question fed as additional input. We hypothesize that a large-scale pretrained transformer model on a massive corpus can understand the meaning of the augmented question and thus successfully answer whether a directed sentiment exists in a text.

Note that our question-answering setup is different from standard question-answering tasks in NLP, as represented by well-known benchmark data such as SQuAD~\cite{rajpurkar2016squad} and WikiQA~\cite{yang2015wikiqa}. Given a question and reference text, the standard task aims at generating answers in a natural language form. In contrast, the question-answering process in DSE2QA requires a binary answer, which can be seen as a special type of question-answering.

\subsubsection{Data Augmentation for DSE2QA}

For each pair of label $l$ and sentence $s$ where the two target entities $p$ and $q$ are masked as \textsc{[Ent1]} and \textsc{[Ent2]} respectively, we augment the training data by transforming the original input into the five tuples using the same sentence and different questions: $t_i$: ($s$, $q_i$, $l_i$) where $i$ is the index of the target relation class: neutral (0), $p\rightarrow q$ with positive sentiment (1), $p\leftarrow q$ with positive sentiment (2), $p\rightarrow q$ with negative sentiment (3), and $p\leftarrow q$ with negative sentiment (4). $l_i$ becomes 1 if $l$ is $i$; otherwise $l_i$ is 0.

We design the auxiliary question $q_{i}$ asking whether the given sentence $s$ is classified as the target sentiment $i$. The list of questions are presented in Table~\ref{tab:aux_question}. For example, $q_1$ asks a model whether a given sentence $s$ contains positive sentiment from $p$ to $q$. Here, we define two types of questions: complete and pseudo. Complete questions are written in a natural language, and pseudo questions only contain keywords that is sufficient to characterize a sentiment class $i$ while ignoring the syntactic structure.

\subsubsection{Model Prediction}

We utilize the BERT-like transformer model~\cite{devlin2018bert}, which can take sentence pairs as input, for making a binary prediction on a given sentence $s$ and question $q_i$. In particular, the model takes 
\begin{quote}
 \textsc{[CLS]} $s$ \textsc{[SEP]} $q_i$ \textsc{[SEP]}   
\end{quote}
as input\footnote{`$<$\textsc{s}$>$ $s$ $<$\textsc{/s}$>$ $<$\textsc{/s}$>$ $q_i$ $<$\textsc{/s}$>$ $<$\textsc{/s}$>$' in RoBERTa} and predicts a value $y_i$ from 0 to 1 that indicates the confidence on the target label $i$. 

\subsubsection{Training and Inference}

For the augmented input of $t_i$, a pretrained BERT-like transformer is trained to predict 1 for $t_l$ and 0 for $t_{i\neq l}$ through the \textsc{[CLS]} representation at the last layer of the transformer model followed by a classification layer. For inference, we made a prediction corresponding to $s$ by $\argmax_i y_i$ where $y_i$ is the prediction outcome of $t_i$. The $y_i$ indicates the confidence on each sentiment $i$, and therefore we take the class of which the value is maximum. 

In the experiments, we utilize the RoBERTa base model for the backbone of DSE2QA and train the model to minimize the binary cross-entropy loss. This approach is different from the RoBERTa classification model that only employs a single sentence input.

\begin{table*}[ht]
\centering
\begin{tabular}{|l|rrr|}
\hline Method & Micro F1 & Macro F1 & $mAP$\\ 
\hline
DSE2QA (Pseudo)       & \textbf{0.7973} & \textbf{0.6766} & \textbf{0.7488}\\
DSE2QA (Complete)    & 0.7726 & 0.6617 & 0.7387\\\hline
RoBERTa    & 0.7486 & 0.6409 & 0.7319\\
LNZ (Combined)                    & 0.7055 & 0.5358 & 0.5295\\
LNZ (Context)                     & 0.6371 & 0.4665 & 0.4921\\
LNZ (EntityPrior)                 & 0.5853 & 0.4063 & 0.414\\\hline
\end{tabular}
\caption{Evaluated performance on the test set. Top performance for each metric is marked as bold.}
\label{table:exp-result}
\end{table*}

\begin{table*}[ht]
\centering
\begin{tabular}{|l|rrrrr|}
\hline Method & 0 & 1 & 2 & 3 & 4\\ 
\hline
DSE2QA (Pseudo)     & \textbf{0.855} & \textbf{0.6519} & \textbf{0.5672} & 0.7402 & \textbf{0.5686}\\
DSE2QA (Complete)   & 0.8293 & 0.6421 & 0.5672 & \textbf{0.7416} & 0.5283\\\hline
RoBERTa             & 0.8054 & 0.6373 & 0.5079 & 0.7184 & 0.5354\\
LNZ (Combined)            & 0.7981 & 0.443 & 0.3333 & 0.5827 & 0.5217\\
LNZ (Context)             & 0.7469 & 0.4069 & 0.2817 & 0.5007 & 0.3964\\
LNZ (EntityPrior)         & 0.7133 & 0.2629 & 0.2353 & 0.4533 & 0.3667\\\hline
\end{tabular}
\caption{F1-score per class measured on the test set. Top performance for each metric is marked as bold.}
\label{table:exp-result-per-class}
\end{table*}

\section{Performance Evaluation}

We evaluate the performance of the proposed DSE2QA approach using our annotated corpus. We compare our method with the current state-of-the-art methods for directional blame detection proposed by~\citet{liang2019blames} (LNZ) as well as a classification model fine-tuned on a pretrained transformer (RoBERTa).

\subsection{Experiment Setups}
For the LNZ models~\cite{liang2019blames}, we set the vocabulary size as 10000. We set the word embedding size, LSTM hidden dimension, and the fully connected layer dimension as 256, 512, and 1024. The search space for the dropout rate is $[0.1, 0.5]$. We train the LNZ models using Adam optimizer with a learning rate of 1e-3~\cite{kingma2014adam}. We adopt an early stopping strategy with the patience of 5. For training RoBERTa and DSE2QA, we followed the procedure of \citet{liu2019roberta} using AdamW~\cite{loshchilov2017decoupled}. We optimize hyperparameters by randomly choosing ten sets for each model and selecting the model with the best performance on the validation set. The learning rate is set to 2e-5 with the epsilon as 1e-6. The weight decay is set to 0.1. 
We apply random oversampling to the training set to make a balanced dataset against the label. We run the experiment five times with different random seeds and report the average scores.

\subsection{Evaluation Results}

We utilize three measures for evaluation: micro-f1, macro-f1, and mean average precision (mAP). Micro-f1 is calculated by (\#correct)/(\#total), which corresponds to the multi-class classification accuracy. Macro-f1 measures an f1-score for each class and averages them with equal importance; therefore, macro-f1 is a more robust measure to a skewed class distribution, such as our annotation data (see Table~\ref{table:data-stat}). Similarly, mAP measures the unweighted average of average precision (AP) on each class; AP summarizes a precision-recall curve varying prediction threshold for a target class.

In Table~\ref{table:exp-result}, we make three observations. First, among classification approaches (the bottom four rows), RoBERTa outperforms the other approaches across the three measures (0.7486/0.6409/0.7319). The LNZ combined model achieves a fair micro-f1 score of 0.7055 but low scores of macro-f1 (0.5358) and mAP(0.5295). This difference is because the combined model (and other non-transformer models) is poor at classifying non-neutral sentiment, which we will further investigate in Table~\ref{table:exp-result-per-class}. Second, DSE2QA with complete questions outperforms RoBERTa with a margin of 0.024 by micro F1. The proposed approach also achieves better performance in macro F1 and mAP. Third, the performance of DSE2QA gets further increased with the usage of pseudo questions, up to the micro-f1 score of 0.7973. This observation implies that a BERT-like transformer model may not need a full sentence to utilize the auxiliary input because it also performs well using fewer keywords for the detection task with an augmented input. 

Table~\ref{table:exp-result-per-class} presents the f1-score measured for each class: neutral (0), positive from the left entity to the right (1), positive from the right to the left (2), negative from the left to the right (3), and negative from the right to the left (4). Here, we make three observations. First, all models perform the best at identifying neutral sentiment (0) compared to the other sentiment classes. Second, non-transformer models (the bottom three rows) are poor at extracting non-neutral sentiment regardless of their direction, which contributes to the decreased macro F1 in Table~\ref{table:exp-result}. Third, among the sentiment classes from the left entity to the right entity (1, 3), transformer models better detect negative sentiment than positive sentiment. The finding suggests that positive entity relationships are more difficult to be captured in news articles, which calls for future studies for a better understanding and model improvement. AP per each class also shows a similar trend, as presented in Table~\ref{table:exp-result-ap-per-class}.

In summary, the proposed approach of solving the directed sentiment extraction task by multiple question-answering tasks outperforms the state-of-the-art classification approaches. The high performance suggests that the model can understand the meaning of augmented questions to some extent, which may build on the language understanding ability of the pretrained RoBERTa.

\section{Analyzing Entity Relationships in News Articles}\label{sec:application}

To demonstrate the utility of the proposed dataset and model, we conduct two case studies to analyze entity-to-entity sentiment relationships presented in recent news articles on political issues: the 2016 U.S. presidential election and the COVID-19 pandemic. For the analyses, we utilize the DSE2QA model with pseudo questions trained on the annotated corpus, and we confirm that the target news articles are not overlapped with the whole set.

\subsection{Case Study 1: 2016 U.S. election}\label{analysis-entity-based}
We study how news media covered the entity relation during the 2016 United States presidential election using a public dataset on news articles between Feb. 2016 to Feb. 2017\footnote{https://www.kaggle.com/snapcrack/all-the-news}. This dataset consists of about 140K news articles in English from fifteen media companies, including CNN, New York Times, and Guardian. We randomly select 3K articles from each month, 39K articles in total. Then we apply the proposed model to all sentences that contain at least two entities from the top-30 most frequent entities, including Donald Trump and Hillary Clinton. 

\begin{figure}[ht]
\centering
    \includegraphics[width=0.40\textwidth]{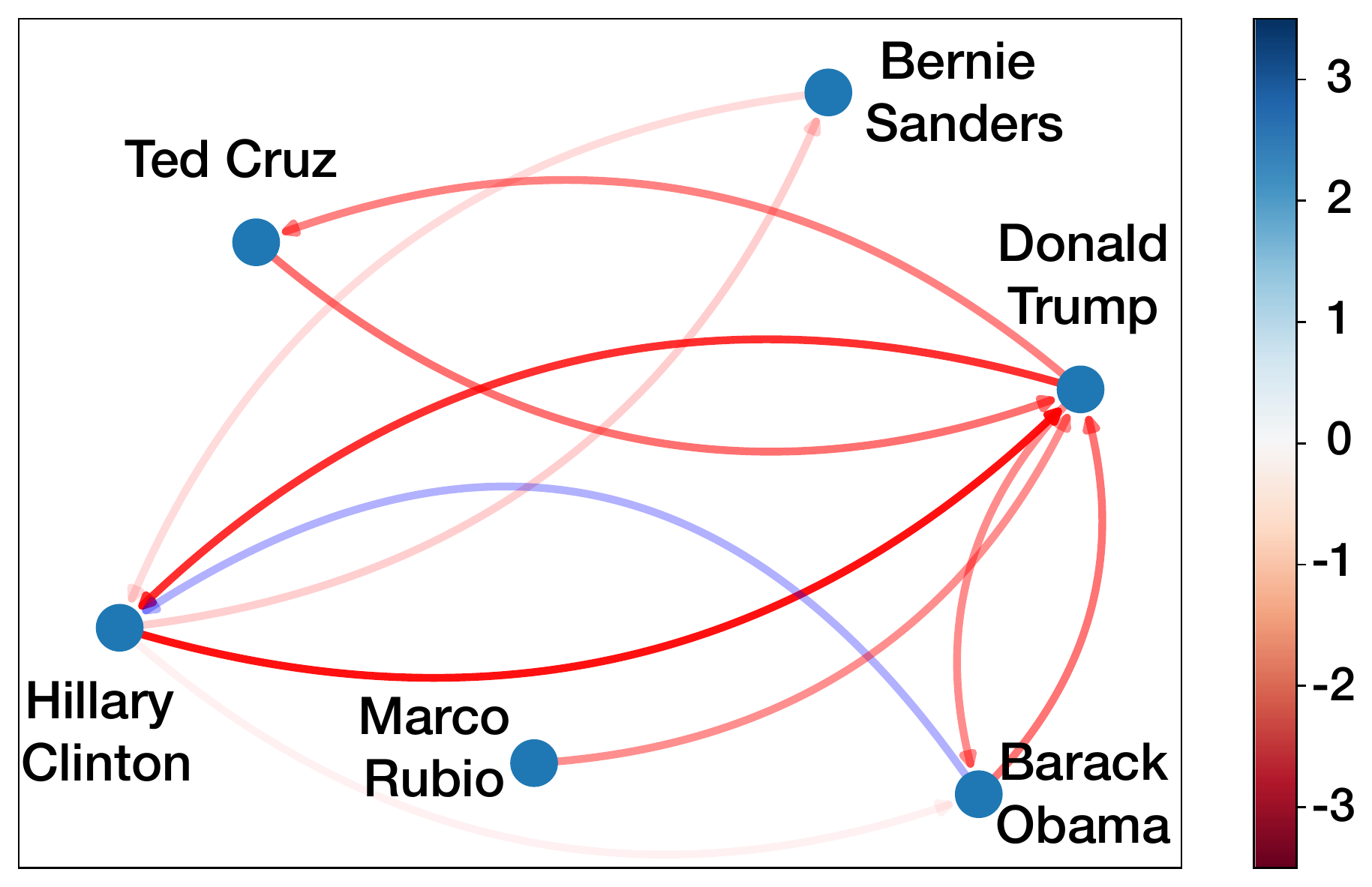}
    \vspace{-5pt}
    \caption{Log ratios of positive (blue) and negative (red) sentiments in directed political relationships in the news articles.}
    \vspace{-5pt}
    \label{fig-graph}
\end{figure}

Figure~\ref{fig-graph} presents the frequently mentioned pairs of politicians. The red color indicates an entity pair tends to contain negative sentiment more than positive, and blue indicates the pairs with the negative sentiment more. The opacity represents the strength of the opinion measured by the log ratio of inferred pairs of positive and negative sentiment. For brevity, we present relations that appear at least 20 times in any sentiment. 
Overall, there are 4.68 times more negative sentiments found than positive ones, which may be explained due to the negative nature of political media~\cite{soroka2014negativity}. 
An interesting observation is the asymmetric relation between Hillary Clinton and Barack Obama. Clinton holds a generally slightly negative opinion towards Obama, while Obama holds a stronger positive opinion towards Clinton. 
Note that our model does not just memorize the entity relationship in the training dataset and apply it to the target dataset as we replace detected entities with tokens in input sentences.  

\subsection{Case Study 2: the COVID-19 pandemic}
In the second study, we investigate how news media portray political entities and their relationships differently according to their political orientations. For example, a right-leaning news outlet may show more negative opinions expressed toward Democrats. To that end, we focused on the recent issue of the COVID-19 pandemic to examine the media bias. We expect the general sentiment about COVID-19 is negative but would like to investigate who blames whom because the messages will have very different meanings according to the sources and targets, differentiated by our data and method.

\begin{table}[t]
\centering
\scalebox{0.75}{
\begin{tabular}{|ccc|}
\hline
Left & Center & Right \\\hline
ABC News  & Associated Press & Breitbart News\\
BuzzFeed News  & Forbes & Daily Mail\\
CBS News  & NPR & Fox News\\
CNN  & Reuter & National Review\\
Democracy Now & USA TODAY & New York Post\\
HuffPost &  & Reason\\
MSNBC & & The American Spectator\\
New York Times & & theblaze.com\\
Slate & & The Daily Caller\\
The Atlantic & & The Daily Wire\\
The Guardian & & The Epoch Times\\
The New Yorker & & The Federalist\\
Time Magazine & & Washington Times\\
Vox & & Wall Street Journal\\
Washington Post & & \\\hline
\end{tabular}
}
\caption{The list target media outlets sorted by alphabetical order.}
\label{table:media_list}
\end{table}

\begin{table}[t]
    \centering
    \scalebox{0.9}{
    \begin{tabular}{|c|ccc|}
    \hline
         Entity & \makecell{Rank\\(Left)} & \makecell{Rank\\(Center)} &  \makecell{Rank\\(Right)}\\\hline
         Donald Trump & 1 & 2 & 1\\
         Republican & 2 & 6 & 6\\
         U.S. & 3 & 1 & 3\\
         Democrat & 4 & 5 & 4\\
         Russia & 5 & 6 & 9\\
         American	 & 6 & 15 & 10\\
         China & 7 & 4 & 5\\
         Obama & 8 & 10 & 7\\
         Chinese & 9 & 3 & 2\\
         Joe Biden & 9 & 8& 7\\\hline
         
    \end{tabular}
    }
    \caption{Frequency rank of top-10 frequent entities targeted with a negative sentiment through an entity-to-entity relationship in the COVID-19 news dataset. The order is sorted by the overall rank in the dataset. }
    \label{tab:negative_target}
\end{table}

To collect a recent news article set, we first compiled a list of 35 popular news media outlets that cover American politics in English. We also ensure the list of news outlets was balanced against political bias, according to the media bias ratings in \url{allsides.com}. For brevity, we consolidate `Lean left' and `Left' into `Left' and `Lean right' and `Right' into `Right.' Table~\ref{table:media_list} presents the list of the target media.

\begin{table*}[t]
    \centering
    \scalebox{0.8}{
    \begin{tabular}{|l|ccc||l|ccc|}
    \multicolumn{4}{c}{\textbf{Frequent pairs in the left-leaning media}} & \multicolumn{4}{c}{\textbf{Frequent pairs in the right-leaning media}}\\\hline
         Entity pairs & \makecell{Rank\\(Left)} & \makecell{Rank\\(Center)} &  \makecell{Rank\\(Right)} & 
         Entity pairs & \makecell{Rank\\(Left)} & \makecell{Rank\\(Center)} &  \makecell{Rank\\(Right)}\\\hline
         Democrat$\rightarrow$ Donald Trump & 1 & 8 & 1 & 	Democrat$\rightarrow$Donald Trump & 1 & 8 & 1\\
         Republican$\rightarrow$Donald Trump & 2 & 66 & 4 & 
         Donald Trump$\rightarrow$Democrat	& 4 & 3 & 2\\
         Twitter$\rightarrow$Donald Trump & 3 & 67& 392 & 
         Donald Trump$\rightarrow$Chinese	 & 30 & 17 & 3\\
         
         Donald Trump$\rightarrow$Democrat	 & 4 & 3& 2 & 
         Republican$\rightarrow$Donald Trump	 & 2 & 66 & 4\\ 
         Bernie Sander$\rightarrow$Donald Trump	 & 5 & 67& 153 & Donald Trump$\rightarrow$China	 & 9 & 10 & 5\\
         Donald Trump$\rightarrow$Ted Cruz	  & 6 & 67& 393 & 
         Donald Trump$\rightarrow$Joe Biden	& 20 & 8 & 6\\
         Republican$\rightarrow$Democrat	 & 7 & 66& 47 & 
         Democrat$\rightarrow$Republican	& 8 & 41 & 7\\
         Democrat$\rightarrow$Republican	& 8 & 41& 7 & 
         Joe Biden$\rightarrow$Donald Trump & 38 & 5 & 8\\
         Donald Trump$\rightarrow$China		 & 9 & 10& 5 & 
         Donald Trump$\rightarrow$Russia	& 23 & 17 & 9\\
         Donald Trump$\rightarrow$Bush & 10 & 67 & 393 & 
         House$\rightarrow$Donald Trump & 12 & 67 & 10\\\hline
         
    \end{tabular}
    }
    \caption{Frequency rank of entity pairs presented with a negative sentiment in the COVID-19 news dataset}
    \label{tab:negative_pair}
\end{table*}

For each of the target media outlets, we collected news articles shared throughout 2020 until September from the Common Crawl corpus, which has been collecting web data since 2008\footnote{https://commoncrawl.org/}. The total number of retrieved news pages are is 256,081; on average, we have 7,113 published news articles for each outlet. 

Next, we selected documents containing at least one of the keywords relevant to COVID-19 by following similar practices used for collecting a Twitter dataset~\cite{chen2020covid}: \textit{coronavirus}, \textit{covid-19}, \textit{COVID19}, and \textit{corona virus}. We consider sentences containing two or more entities for the target of inference, and every relationship pair is inferred when there are more than two entities in a sentence. The final set consists of 6,180 articles involving 1,078,377 entity pairs for COVID-19.

Table~\ref{tab:negative_target} presents the list of 10 frequent entities that are manifested through entity-to-entity relationships with a negative sentiment. While Donald Trump was the most frequent target of blames in the total data, the results show that the right-leaning media tend to express a negative sentiment toward China (\#5) and Chinese people (\#2) more frequently. To examine the difference systematically, we measure the spearman rank correlation coefficient for the whole list of entities that appeared in the dataset. The rank in the left-leaning media and that in right-leaning media exhibits a highly negative correlation of -0.5722 (\textit{p}$<$0.001), which suggests that the list of political entities presented with a negative sentiment significantly varies across news media according to their political orientation. Such a high level of negative correlation is also observed in the ranks for the source entity in negative relationships ($-$0.4129 with \textit{p}$<$0.001) and the source/target entities in positive relationships ($-$0.5605/$-$0.7929 with \textit{p}$<$0.001).

Going further, we analyze the differences between frequently presented entity pairs with negative sentiment by the left and right-leaning media, respectively. Table~\ref{tab:negative_pair} presents the rank of each entity pair in the media groups according to their political orientation. In the left-leaning media of our dataset, Donald Trump appears as either source or target in the top-10 frequent negative pairs except for the pair of Republican$\rightarrow$Democrat and vice versa. On the contrary, the top-10 pairs in the right-leaning media include the international relationships of Donald Trump to the other countries. In other words, the left-leaning media may try to frame COVID-19 as a \textbf{domestic event} focusing on how the President handles it and how people respond to his crisis management, and the right-leaning media focus more on \textbf{foreign policies and international relationship} especially between the U.S. and China. 
For the whole set of entity pairs, the rank correlation between the left and right media is $-$0.4847 (\textit{p}$<$0.001) for negative sentiment and $-$0.7929 (\textit{p}$<$0.001) for positive sentiment. These negative correlations imply that the news media has a bias in selecting issues to cover (selection bias) and presenting relationships of political entities (presentation bias). 

\section{Conclusion}\label{sec:conclusion}
Detecting who blames or endorses whom in news articles is a critical ability in understanding opinions and relationship between political actors in news media. This paper provides a computational tool based on natural language processing for facilitating interdisciplinary studies using text in news and social media.

We introduced a new problem of identifying directed sentiment relationships between political entities, called directed sentiment extraction. We constructed a training corpus of which entity relationship is manually annotated for each sentence. This dataset can serve as a benchmark for future studies. A potential future direction is to build an improved version of the dataset where sentiment relationships between political entities appear across multiple sentences in news articles. Our problem setup is similar to the document-level sentiment inference task of \citet{choi-etal-2016-document}; they infer entity-to-entity sentiment in a sentence among three classes (positive, unbiased, negative), and the sentence-level outcomes are merged to infer a 5-class sentiment for a document. Future studies could extend our sentence-level approach to document-level sentiment inference.

To tackle the problem, we proposed DSE2QA (Directed Sentiment Extraction to Question-Answering), which is a simple yet effective method of utilizing BERT-like pretrained transformers by predicting answers for binary questions on whether a sentiment relationship exists in a given text. Answers for each sentiment class are aggregated to make a final guess. Evaluation experiments show the approach outperforms state-of-the-art classification models, such as the fine-tuned RoBERTa classification model. We hypothesize the language understanding ability of the BERT-like pretrained transformer may contribute to the high performance, combined with its facility of taking auxiliary input. Furthermore, the performance increase with the pseudo questions implies that a few keywords may suffice to make an inquiry. Future research could investigate which kind of pretrained transformer is the most effective for understanding the meaning of the augmented question, as DSE2QA's backbone can be replaced with any BERT-like transformer. Also, this study calls for future studies on advanced methods for directed sentiment extraction. 
A potential approach could jointly learn entity recognition and directed sentiment extraction as similarly tackled by a study on information extraction~\cite{bekoulis2018joint}. 

As the last step, we conducted case studies by analyzing directed sentiments in news text for the US election and COVID-19 pandemic. The observations not only add empirical understandings to the social science research but also highlight the utility of the proposed problem, dataset, and model for political news analysis. We believe the proposed method can therefore further the current interdisciplinary efforts of the NLP, machine learning, and the social science communities~\cite{grimmer2013text,roberts2014structural,joo2018image}.

\section*{Acknowledgements}
We thank anonymous reviewers for their valuable comments. 
This work was supported by NSF SBE/SMA \#1831848 ``RIDIR: Integrated Communication Database and Computational Tools''.

\section*{Ethics and Impact Statement}
In online news and social media, people express diverse sentiment toward a target through text, such as blame, support, endorsement, to list a few. Quantifying and understanding the patterns is of significant interest in social science, but the lack of automated methods makes it difficult to handle large-scale data, which can reveal patterns in a comprehensive view. In this light, this study aims to develop automated methods of identifying directed sentiment between entities by defining a new NLP problem: directed sentiment extraction. The newly annotated dataset will facilitate future development of the NLP methods, and the DSE2QA approach will serve as a strong baseline. 

The development of an automated method will have a broader impact by tackling real-world challenges such as bias in news reporting against political orientation, with potential collaboration with social science. Moreover, it will enable the discovery of hidden biases with regard to sentiment in online text, which can be mistakenly learned through data-driven methods. A fine-grained understanding of sentiment relationships will broadly contribute to building a fair machine learning model, which is of significant interest in AI ethics.

{
\bibliography{acl2021}
\bibliographystyle{acl_natbib}
}

\clearpage

\appendix
\setcounter{table}{0}
\renewcommand{\thetable}{A\arabic{table}}
\setcounter{figure}{0}
\renewcommand{\thefigure}{A\arabic{figure}}

\begin{table*}[ht]
    \centering
    \small
    \begin{tabular}{|l|}
    \hline
    
         \textbf{Instruction:}\\
         Based the given sentence, please identify if one entity (person, organization, or country) holds a positive or \\negative opinion towards another entity. 
         
         There will be two entities in total. One is in \textcolor{red}{red} and the other is in \textcolor{blue}{blue}.\\
        ~\\
        It is also possible neither positive nor negative opinion exists in the sentence. \\Please annotate such cases as neutral.\\
        ~\\
        Please classify the sentence based on what people say instead of what they do. \\For example, if a sentence only states the police arrest someone, this sentence should be classified as neutral. \\Instead, if the police accuses someone of commiting a crime, this sentence should be classified as police holds \\a negative opinion towards the person.\\
        ~\\
        \textbf{Examples:}\\
        - Earlier on Tuesday, Mr. Trump criticized General Motors for making cars in Mexico.\\  (Negative: Trump holds a negative opinion towards General Motors)\\
        - Hugo Ras has been accused of killing other people’s rhino, and for that South Africans condemn him. \\ (Negative: South Africans holds a negative opinion towards Hugo Ras)\\
        - DAVID Cameron's accused the Conservatives of failing to devolve essential welfare powers to, as agreed \\by the cross-party Smith Commission which considered further powers for the Scottish parliament last year. \\ (Neutral: There is no direct opinions between these two entities.)\\
        - Prime Minister Stephen Harper shakes hands with Petty Harbour, N.L., during a campaign event in Toronto \\on Sept. 18, 2015. (Neutral: They shake hands just for politeness. No opinions exist.)\\
        - Obama pulled Clinton into his administration after he defeated her in 2008 primary and has effusively praised \\her tenure as secretary of state. (Positive. Obama holds a positive opinion towards Clinton.)\\
        - Earlier on Tuesday, Mr. Trump has not comments on General Motors making cars in Mexico. (Neutral.)\\
        ~\\
        \textbf{Note:}\\
        There are multiple annotators for each sentence. Your response will be judged as failed when it is different with\\other annotators. If the percentage of failed response from one annotator is above a threshold, the annotator \\will NOT get paid for ALL responses. Thanks for your participation.
        \\\hline
    \end{tabular}
    \caption{Instruction used for educating annotators in Amazon Mechanical Turk. }
    \label{tab:instruction}
\end{table*}

\begin{table*}[ht]
\centering
\scalebox{1}{
\begin{tabular}{|l|rr|rrr|}
\hline Method & \makecell{Num. \\parameters} & \makecell{Avg. runtime\\per epoch} & Micro F1 & Macro F1 & $mAP$\\ 
\hline
DSE2QA (Pseudo)       &  125M + 1536 & 2154s & \textbf{0.8072} & \textbf{0.6827} & \textbf{0.7528} \\
DSE2QA (Complete)    & 125M + 1536 & 2149s & 0.7892 & 0.6751 & 0.7724\\\hline
RoBERTa    &  125M + 3840 & 437s & 0.7618 & 0.6516 & 0.7493\\
LNZ (Combined)                    & 3.03M & 12s & 0.694 & 0.5189 & 0.4819\\
LNZ (Context)                     & 2.9M & 12s & 0.6331 & 0.4518 & 0.3908\\
LNZ (EntityPrior)                 & 2.65M & 4.8s & 0.5914 & 0.4427 & 0.31\\\hline
\end{tabular}}
\caption{Model details and evaluated performance on the validation set. Top performance for each metric is marked as bold.}
\label{table:validation_performance}
\end{table*}

\begin{table*}[ht]
\centering
\scalebox{1}{
\begin{tabular}{|l|rrrrr|}
\hline Method & 0 & 1 & 2 & 3 & 4\\ 
\hline
DSE2QA (Pseudo)     & 0.9316 & 0.7157 & \textbf{0.5952} & 0.8358 & \textbf{0.6658}\\
DSE2QA (Complete)   & \textbf{0.9341} & \textbf{0.7228} & 0.5747 & \textbf{0.8457} & 0.6161\\\hline
RoBERTa             & 0.929 & 0.7299 & 0.5236 & 0.8232 & 0.6536\\
LNZ (Combined)            & 0.8452 & 0.4554 & 0.2887 & 0.6273 & 0.4311\\
LNZ (Context)             & 0.8233 & 0.4261 & 0.2887 & 0.568 & 0.3545\\
LNZ (EntityPrior)         & 0.7834 & 0.2181 & 0.2248 & 0.4405 & 0.4033\\\hline
\end{tabular}}
\caption{AP per class measured on the test set. Top performance for each metric is marked as bold.}
\label{table:exp-result-ap-per-class}
\end{table*}

\end{document}